\documentclass{article}

\usepackage{microtype}
\usepackage{graphicx}
\usepackage{subfigure}
\usepackage{booktabs} 
\usepackage{algorithm}
\usepackage{algorithmic}
\usepackage{xcolor}
\usepackage{listings}

\usepackage{hyperref}

\usepackage[accepted]{icml2024}

\usepackage{amsmath}
\usepackage{amssymb}
\usepackage{mathtools}
\usepackage{amsthm}
\usepackage{xcolor,colortbl}

\usepackage[capitalize,noabbrev]{cleveref}

\theoremstyle{plain}

\theoremstyle{definition}

\theoremstyle{remark}

\definecolor{mygray}{gray}{0.6}
 
\definecolor{lightgray}{gray}{0}

\usepackage[textsize=tiny]{todonotes}

\icmltitlerunning{SE(3)-Hyena Operator}

\begin{document}

\twocolumn[
\icmltitle{SE(3)-Hyena Operator for Scalable Equivariant Learning}




\begin{icmlauthorlist}
\icmlauthor{Artem Moskalev}{jnj}
\icmlauthor{Mangal Prakash}{jnj}
\icmlauthor{Rui Liao}{jnj}
\icmlauthor{Tommaso Mansi}{jnj}
\end{icmlauthorlist}

\icmlcorrespondingauthor{Artem Moskalev}{ammoskalevartem@gmail.com}

\icmlaffiliation{jnj}{Johnson and Johnson Innovative Medicine}

\icmlkeywords{Equivariance, Global context, Scalability, Long convolution, ICML}

\vskip 0.3in
]
\printAffiliationsAndNotice{}




\begin{abstract}

Modeling global geometric context while maintaining equivariance is crucial for accurate predictions in many fields such as biology, chemistry, or vision. Yet, this is challenging due to the computational demands of processing high-dimensional data at scale. Existing approaches such as equivariant self-attention or distance-based message passing, suffer from quadratic complexity with respect to sequence length, while localized methods sacrifice global information. Inspired by the recent success of state-space and long-convolutional models, in this work, we introduce SE(3)-Hyena operator, an equivariant long-convolutional model based on the Hyena operator. The SE(3)-Hyena captures global geometric context at sub-quadratic complexity while maintaining equivariance to rotations and translations. Evaluated on equivariant associative recall and n-body modeling, SE(3)-Hyena matches or outperforms equivariant self-attention while requiring significantly less memory and computational resources for long sequences. Our model processes the geometric context of $20k$ tokens $\times3.5$ faster than the equivariant transformer and allows $\times175$ longer a context within the same memory budget. 
\end{abstract}

\section{Introduction}

Modeling global geometric context while preserving equivariance is crucial in many real-world tasks. The properties of a protein depend on the global interaction of its residues \cite{baker2001ProteinSP}. Similarly, the global geometry of DNA and RNA dictates their functional properties \cite{leontis2001geometric,sato2021rna}. In computer vision, modeling global geometric context is crucial when working with point clouds or meshes \cite{thomas2018tensor,de2020gauge}. In all these tasks, maintaining equivariance while capturing global context is essential for robust modeling and prediction.

Processing global geometric context with equivariance is challenging due to the computational demands of processing high-dimensional data at scale. Existing methods either rely on global all-to-all operators such as self-attention \cite{liao2023equiformer,dehaan2023euclidean,brehmer2023geometric}, which do not scale well due to their quadratic complexity, or they restrict processing to local neighborhoods \cite{thomas2018tensor,kohler2020equivariant,fuchs2020se}, losing valuable global information. This limitation is a significant practical bottleneck, necessitating more efficient solutions for scalable equivariant modeling with a global geometric context.

\begin{figure}[t!]
  \centering
    \includegraphics[width=0.95\linewidth]{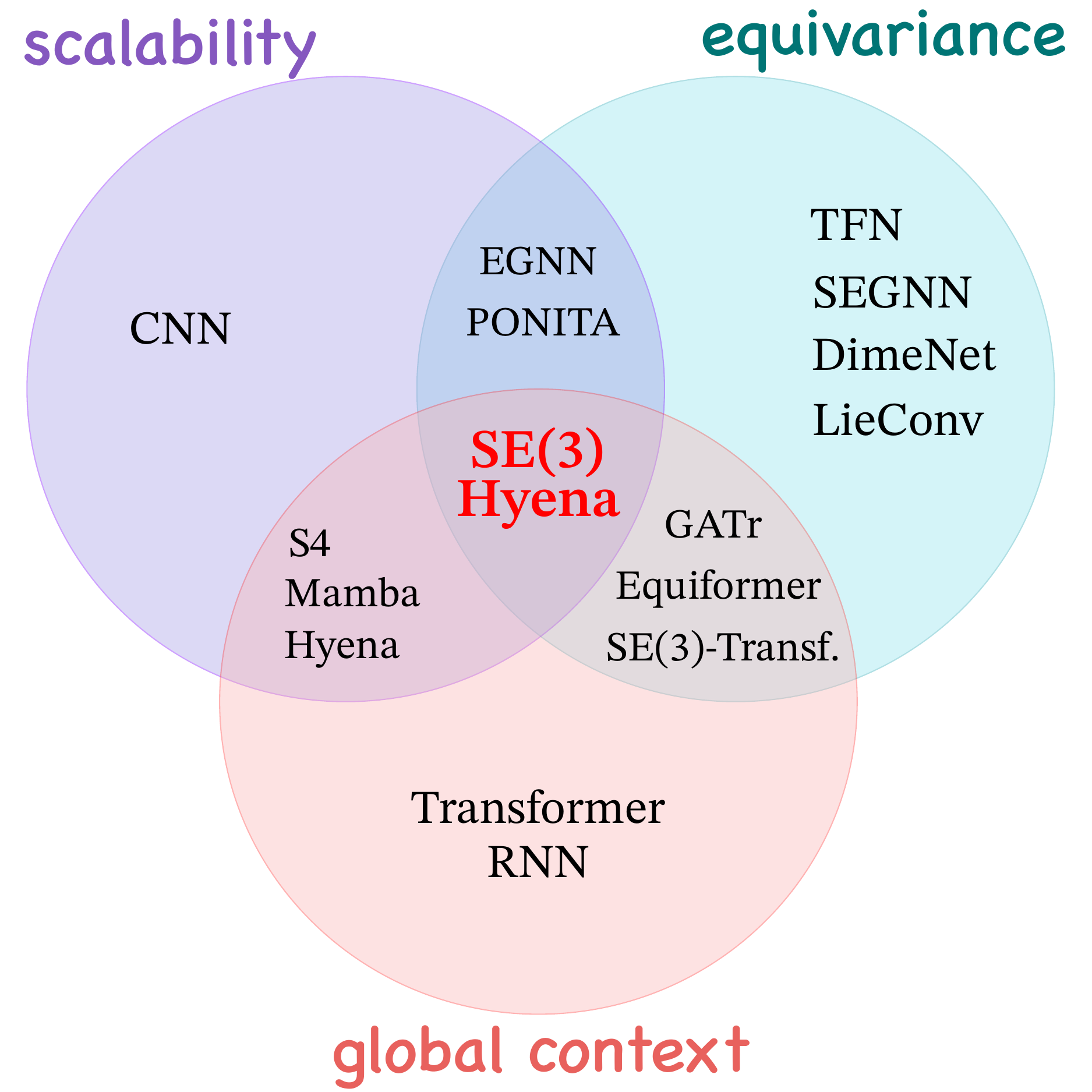}
    \caption{
    \textbf{SE(3)-Hyena operator marries global context, equivariance and scalability towards long sequences.}
    SE(3)-Hyena operator can process global geometric context in sub-quadratic time while preserving equivariance to rotations and translations.}
  \label{fig:teaser}
\end{figure}

An efficient algorithm for modeling global context should support parallelization during training while maintaining bounded computational costs relative to sequence length during inference. One approach involves recurrent operators \cite{orvieto2023resurrecting,de2024griffin}, which provide bounded compute but lack easy parallelization. Another family of methods relies on self-attention \cite{vaswani2017attention} allowing parallel processing at the cost of quadratic computational complexity. The most recent advances leverage state-space \cite{gu2021combining,fu2022hungry,gu2023mamba} and long-convolutional \cite{romero2021ckconv,poli2023hyena} frameworks, enabling global context reasoning in sub-quadratic time with easy parallelization. Extending these models to accommodate equivariance remains an unexplored direction.

Inspired by the success of state-space and long-convolutional methods, in this work we propose a SE(3)-equivariant long-convolutional model based on the recent Hyena operator~\cite{poli2023hyena}. The SE(3)-Hyena efficiently models global geometric context in sub-quadratic time while preserving equivariance to rotations and translations (Figure~\ref{fig:teaser}). Central to our method is the equivariant vector long convolution that leverages cross products between equivariant queries and keys. This vector convolution can be implemented in the Fourier domain with a computational complexity of $O(N log_2 N)$, enabling scaling to much longer sequences than self-attention. We evaluate the proposed SE(3)-Hyena against its self-attention counterpart on a novel equivariant associative recall benchmark and the n-body dynamical system modeling task. Our results suggest that SE(3)-Hyena matches or surpasses equivariant self-attention in performance while requiring less memory and compute for long sequences. In particular, for a sequence of $20k$ tokens, the equivariant Hyena runs $\times 3.5$ faster than the equivariant transformer approach. Notably, when the equivariant transformer runs out of memory on sequences over $20k$ tokens, \textit{our model can handle up to $3.5M$ million tokens on a single GPU}, providing up to $175$ times longer context length within the same computational budget.

To sum up, we make the following contributions:

\begin{itemize}
    \item We propose SE(3)-equivariant Hyena operator which enables modeling global geometric context in sub-quadratic time.
    \item We propose an equivariant counterpart for the mechanistic interpretability associative recall task.
    \item We demonstrate that the equivariant Hyena matches or outperforms the equivariant transformer, while requiring significantly less memory and compute for long-context modeling.
\end{itemize}
\section{Related work}

\paragraph{Equivariance} Equivariance to group transformations, particularly rotations and translations in 3D, is crucial for modeling physical systems ~\cite{zhang2023artificial}. \citet{schutt2017schnet} condition continuous convolutional filters on relative distances to build model invariant to rotations and translations. \citet{thomas2018tensor,fuchs2020se,brandstetter2021geometric,liao2023equiformer,bekkers2023fast} utilize spherical harmonics as a steerable basis which enables equivariance between higher-order representations. Since computing spherical harmonics can be expensive, \citet{jing2021learning,jing2021equivariant,satorras2021n,deng2021vn} focus on directly updating vector-valued features to maintain equivariance, while \citet{zhdanov2024implicit} employ another equivariant network to implicitly parameterize steerable kernels. Another recent line of work ~\cite{ruhe2023clifford,ruhe2023geometric,brehmer2023geometric,zhdanov2024cliffordsteerable} employs geometric algebra representation which natively provides a flexible framework for processing symmetries in the data~\cite{dorst2009geometric}.

While these works focus on how to build equivariance into a neural network, in this paper we focus on efficient equivariance to model global geometric contexts at scale.

\paragraph{Modeling geometric context} Various strategies are employed to process context information in geometric data. Convolutional methods aggregate context linearly within a local neighborhood, guided by either a graph topology~\cite{kipf2016semi} or spatial relations in geometric graphs~\cite{schutt2017schnet,wu2019pointconv,thomas2018tensor}. Message-passing framework~\cite{gilmer2017neural} generalizes convolutions, facilitating the exchange of nonlinear messages between nodes with learnable message functions. These approaches are favored for their simplicity, balanced computational demands, and expressiveness~\cite{wu2020comprehensive}. However, they are limited to local interactions and are known to suffer from oversmoothing~\cite{rusch2023survey}. This hinders building deep message-passing networks capable of encompassing a global geometric context in a receptive field. To address these limitations, recent methods have turned to self-attention mechanisms for graph~\cite{yang2021graphformers,kreuzer2021rethinking,kim2022pure,rampavsek2022recipe} and geometric graph~\cite{fuchs2020se,liao2023equiformer,brehmer2023geometric} data, outperforming convolutional and message-passing approaches. Yet, the quadratic computational cost of self-attention poses significant challenges when modeling large-scale physical systems. In this work, we aim to develop a method for global geometric context processing with sub-quadratic computational complexity.

\paragraph{State-space and long-convolutional models} 

The quadratic computational complexity of self-attention has driven the exploration of alternatives for modeling long context. Structured state-space models~\cite{gu2021combining} have emerged as a promising alternative, integrating recurrent and convolutional mechanisms within a single framework. These models enable parallelized training in a convolutional mode and maintain linear complexity with respect to sequence length in a recurrent mode. Models like  S4~\cite{gu2021efficiently}, H3~\cite{fu2022hungry}, and Mamba~\cite{gu2023mamba,li2024mamba} have consistently matched or exceeded transformer performance in diverse tasks such as genomics~\cite{schiff2024caduceus}, long-range language~\cite{wang2024state}, and vision tasks~\cite{zhu2024vision}. Concurrently, another line of work integrates long-convolutional framework with implicit filters~\cite{sitzmann2020implicit,romero2021ckconv,zhdanov2024implicit} to capture global sequence context. The implicit filter formulation allows for data-controlled filtering similar to transformers, while FFT-based long convolution enables global context aggregation in sub-quadratic time~\cite{poli2023hyena}. Such models have shown competitive performance comparable to state-space and transformer architectures in time-series modeling~\cite{romero2021ckconv}, genomics~\cite{nguyen2024hyenadna}, and vision tasks~\cite{poli2023hyena}.

Although state-space and long-convolutional methods dramatically reduced the computational costs associated with processing long sequences, their application to geometric data requiring equivariance remains unexplored. In this work, we adapt the recently proposed Hyena operator~\cite{poli2023hyena} to incorporate SE(3) equivariance. To the best of our knowledge, this is the first equivariant long-convolutional model that can process global geometric contexts with sub-quadratic memory and time requirements.

\section{Method}

We consider tasks that require modeling invariant and equivariant features in geometric graphs. A geometric graph of $N$ nodes is represented by a set of features $\{ \mathbf{x}_{i}, \mathbf{f}_{i} \}_{i=1}^{N}$ where $\mathbf{x}_{i} \in \mathbb{R}^{3}$ represents vector features (e.g. coordinates or velocities), and $\mathbf{f}_{i} \in \mathbb{R}^{S}$ represents scalar features (e.g. atom types, charges or fingerprints). We call $\mathbf{x}_{i}$ geometric or vector tokens, and $\mathbf{f}_{i}$ are scalar tokens. When working with geometric graphs, a neural network must respect symmetries of the input space such as rotation or translation. That means, a model must be equivariant with respect to geometric tokens and invariant with respect to scalar tokens.

\begin{figure*}[t!]
  \centering
    \includegraphics[width=0.95\linewidth]{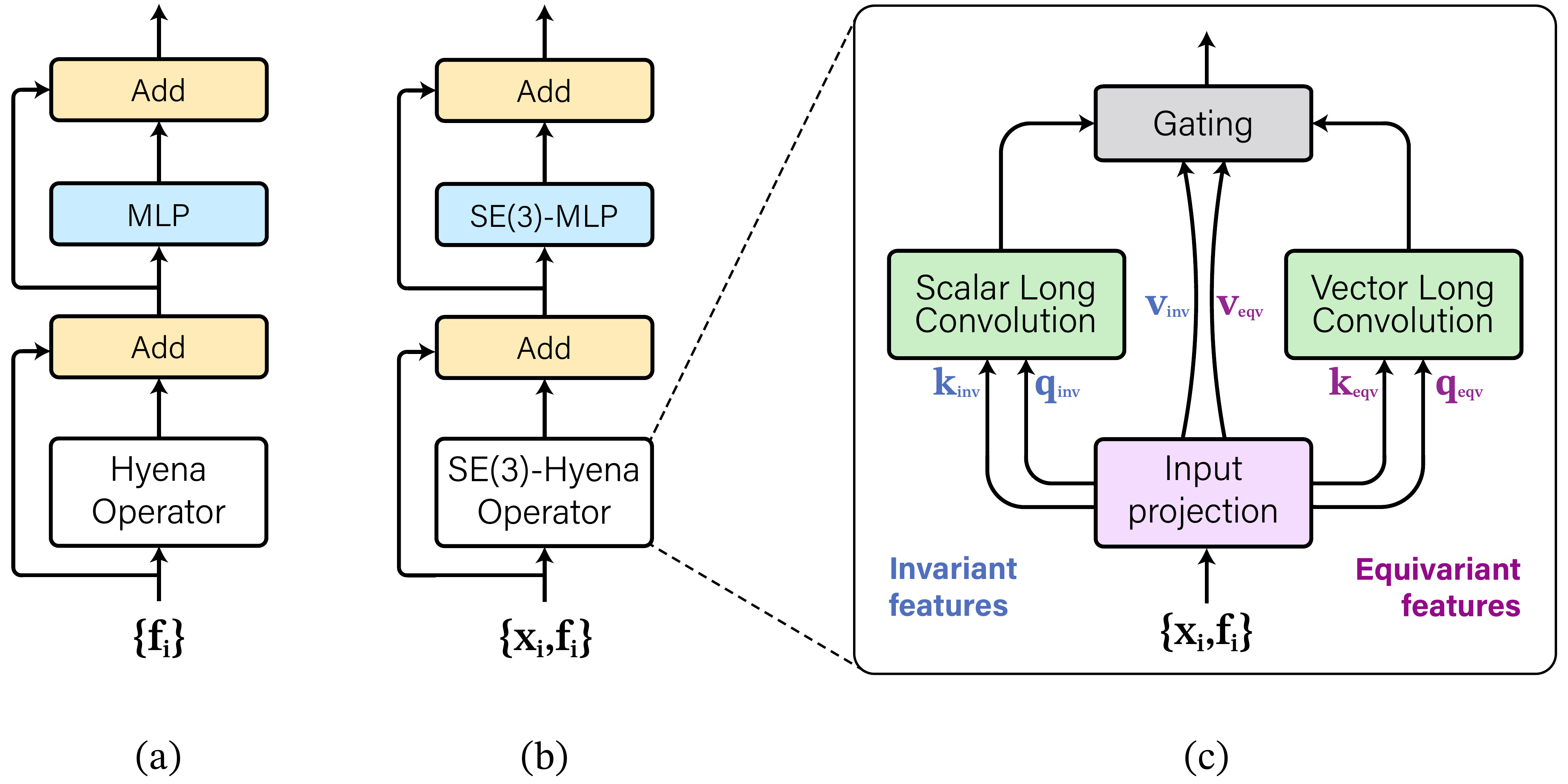}
    \caption{\small 
    \textbf{SE(3)-Hyena building blocks.}
    \textbf{(a) }Schematic of existing Hyena architecture~\cite{poli2023hyena}. \textbf{(b)} The proposed architecture consists of the SE(3)-Hyena operator, residual connections, and an equivariant MLP. \textbf{(c)} The block architecture of SE(3)-Hyena operator consists of two streams processing invariant and equivariant features. The key components are scalar and vector long convolution responsible for global context aggregation.}
  \label{fig:arch}
\end{figure*}

\subsection{SE(3)-Hyena operator}

SE(3)-Hyena operator consists of invariant and equivariant streams which are responsible for processing scalar and vector features respectively, as illustrated in Figure~\ref{fig:arch}. This way, the model takes scalar and vector inputs, and outputs processed scalar and vector features. 

Formally, $\Psi: \mathbb{R}^{3} \times \mathbb{R}^{S} \rightarrow \mathbb{R}^{3} \times \mathbb{R}^{D}$ satisfies equivariance property:

\begin{equation*}
    \label{eq:ghyena}
    \{ L_{g}(\hat{\mathbf{x}}_{i}), \hat{\mathbf{f}}_{i} \}_{i=1}^{N} = \Psi \left( \{ L_{g}(\mathbf{x}_{i}), \mathbf{f}_{i} \}_{i=1}^{N} \right)
\end{equation*}

where $L_{g}: \mathbb{R}^{3} \rightarrow \mathbb{R}^{3}$ is a representation of a group action $g \in SE(3)$. 
Thus, geometric tokens $\mathbf{x}_i$ transform accordingly with the group action while scalar tokens $\mathbf{f}_i$ remain invariant.

For both invariant and equivariant streams, we make the overall information flow similar to the information flow of a transformer with the key difference in how the global context is aggregated. 
Akin to the transformer architecture, the input is firstly projected into keys, queries, and values. 
Next, the global context is aggregated via long convolution and gating. 
Finally, the global context is combined with values, and then projected via a feed-forward network. 

\subsection{SE(3)-Hyena modules}

We design each layer to be equivariant with respect to transformations of geometric tokens, and invariant for scalar tokens. 
This way, a model consisting of a composition of equivariant layers is also equivariant by induction~\cite{weiler2019general}.

\paragraph{Input projection}

Similar to transformers, we firstly project input tokens to queries, keys and values. Because our SE(3)-Hyena operator consists of equivariant and invariant streams, we need to obtain queries, keys and values for both invariant and equivariant features. We define the projection layer 
$\phi:\mathbb{R}^{3} \times \mathbb{R}^{S} \rightarrow (\mathbb{R}^{3})^{3} \times (\mathbb{R}^{D})^{3}$ as $\mathbf{z}^{eqv}_i, \mathbf{z}^{inv}_i = \phi \left(\mathbf{x}_{i}, \mathbf{f}_{i}\right)$ where for $i$-th geometric token $\mathbf{z}^{eqv}_i = [\mathbf{q}^{eqv}_i, \mathbf{k}^{eqv}_i, \mathbf{v}^{eqv}_i]$ is equivariant query, key, and value, each represented by a vector in $\mathbb{R}^{3}$, and $\mathbf{z}^{inv}_i = [\mathbf{q}^{inv}_i, \mathbf{k}^{inv}_i, \mathbf{v}^{inv}_i]$ represents invariant query, key and value of dimensions $\mathbb{R}^{D}$ for $i$-th scalar token. This way, the projection layer emits both scalar and vector query, key, value triplets while also allowing interaction between equivariant and invariant subspaces. 

To simultaneously preserve equivariance for geometric tokens and invariance for scalar tokens, while allowing interaction between them, we adopt E(n)-equivariant Clifford MLP~\cite{ruhe2023clifford} as the input projection function. The inputs are firstly embedded into $CL(\mathbb{R}^3, q)$ Clifford algebra, and then processed via a geometric linear layer, geometric product, and normalization layers. Finally, the output is projected back to scalar and vector features by grade-one and grade-two projections respectively.

\paragraph{Scalar long convolution} 
To allow global context aggregation for invariant scalar features, we rely on long convolution~\cite{romero2021ckconv, poli2023hyena} between query and key tokens. We treat the queries as input signal projection, and the keys constitute a data-controlled implicit filter. Similar to \citet{romero2021ckconv,poli2023hyena}, we employ circular FFT-convolution to reduce the computational complexity. Let $\mathbf{q}^{inv}$ and $\mathbf{k}^{inv}$ be two sequences of length $N$ composed of sets of one-dimensional invariant queries $\{ \mathbf{q}^{inv}_i \}_{i=1}^{N}$ and keys $\{ \mathbf{k}^{inv}_i \}_{i=1}^{N}$ respectively. Then, the global context can be aggregated by FFT-convolution as:


\begin{align}
\label{eq:sca_conv}
    \mathbf{q}^{inv} \circledast \mathbf{k}^{inv} &= \mathbf{F}^H \mathbf{\Lambda_k} \mathbf{F} \mathbf{q}^{inv} \notag \\
    &= \mathbf{F}^H \text{diag}(\mathbf{F}\mathbf{k}^{inv}) \mathbf{F} \mathbf{q}^{inv}
\end{align}

where $\textbf{F}$ is a discrete Fourier transform matrix, and $\text{diag}(\mathbf{F}\mathbf{k}^{inv})$ is a diagonal matrix containing Fourier transform of the kernel $\mathbf{k}^{inv}$.

In the case when query's and key's dimension $D>1$, the scalar FFT-convolution runs separately for each dimension, rendering computational complexity of $O(D N log_2 N)$ that is sub-quadratic in sequence length.

\paragraph{SE(3) vector long convolution}
To allow global context aggregation for geometric tokens, we build equivariant vector long convolution. 
While a scalar convolution relies on dot-products between scalar signals, vector convolution operates with vector cross products $\times$ between vector signals. 
Formally, given a vector signal consisting of $N$ vector tokens $\mathbf{q}^{eqv} \in \mathbb{R}^{N \times 3}$ and a vector kernel $\mathbf{k}^{eqv} \in \mathbb{R}^{N \times 3}$, we define the vector long-convolution as:

\begin{equation}
\label{eq:vec_conv}
    \left( \mathbf{q}^{eqv} \circledast_{\times} \mathbf{k}^{eqv} \right)_{i} = \sum_{j=1}^{N} \mathbf{q}^{eqv}_{i} \times \mathbf{k}^{eqv}_{j - i}
\end{equation}

The computational complexity of a naive implementation for the vector convolution is quadratic since both the signal and a kernel are of the full length. To reduce the computational complexity, we show how the vector convolution can be formulated as a series of scalar convolutions that can be efficiently carried out by the FFT. This is due to the fact that a cross product can be written element-wise through the series of scalar products as $\left(\mathbf{a} \times \mathbf{b}\right)[l] = \mathbf{\varepsilon}_{lhp} \textbf{a}[h] \textbf{b}[p]$ where $\varepsilon$ is \textit{Levi-Civita} symbol, and $\textbf{a}[h]$ denotes a projection onto $h$-th basis vector. Thus, the $l$-th component of the vector convolution in Equation~\ref{eq:vec_conv} can be written element-wise as:


\begin{align}
\label{eq:vec_conv_levi}
    \left( \mathbf{q}^{eqv} \circledast_{\times} \mathbf{k}^{eqv} \right)_{i}[l] &= \varepsilon_{lhp} \sum_{j=1}^{N} \textbf{q}^{eqv}_i[h] \hspace{1mm} \textbf{k}^{eqv}_{j-i}[p] \notag
    \\
    &= \varepsilon_{lhp} \left( \textbf{q}^{eqv}[h] \circledast \textbf{k}^{eqv}[p] \right)_{i} 
\end{align}

Thus, we can obtain $l$-th component of a resulting vector signal via a scalar convolution over the $h$-th and $p$-th components of the sequences $\mathbf{q}^{eqv}$ and $\mathbf{k}^{eqv}$ respectively. Since the scalar convolution can be implemented with the FFT, decomposing the vector convolution to the series of scalar convolutions allows reducing its quadratic complexity to $O(N log_2 N)$.

Since a cross product is already equivariant to rotations, the whole vector convolution is also equivariant to rotations provided the queries $\mathbf{q}^{eqv}$ and the keys $\mathbf{k}^{eqv}$ are rotated accordingly. The latter is guaranteed when the input projection function is equivariant. We can further obtain SE(3) equivariance by firstly centering (by subtracting the center of mass) and then uncentering with respect to translation.

\paragraph{Gating} 
Similar to transformers, we aim to enable data-controlled gating. Similar to the input projection layer, we employ Clifford MLP to obtain a gating mask while allowing interaction between invariant and equivariant subspaces. The output of the Clifford MLP $\gamma: \mathbb{R}^{3} \times \mathbb{R}^{D} \rightarrow (\mathbb{R})^2$ is two grade-one projected scalar features $m^{eqv}_i, m^{inv}_i$ that are passed through a sigmoid function. The sigmoid outputs are multiplied element-wise with the scalar and vector tokens for the invariant and equivariant streams, respectively. In other words, given the input vector token $\mathbf{x}^{eqv}_{i}$, the gated token $\hat{\mathbf{x}}^{eqv}_{i}$ is computed as $\hat{\mathbf{x}}^{eqv}_{i} = \sigma(m^{eqv}_i) \cdot \mathbf{x}^{eqv}_{i}$, with scalar tokens processed in a similar manner.

To align the information flow with that of transformers, the gating is applied on top of the long convolution between queries and keys. Finally, the gated tokens are integrated with value tokens $\mathbf{v}^{eqv}_i$ and $\mathbf{v}^{inv}_i$ using cross and element-wise products for scalar and vector tokens respectively.

By employing grade-one output projection in the Clifford MLP to obtain masking values, the gating mechanism remains E(3)-invariant, thus maintaining invariance for scalar tokens and preserving the equivariance of vector tokens.

\paragraph{Output projection} We add a residual connection between input tokens and the result from the gating layer, and pass it through the output equivariant Clifford MLP $\xi: \mathbb{R}^{3} \times \mathbb{R}^{D} \rightarrow \mathbb{R}^{3} \times \mathbb{R}^{D}$ with grade-one and grade-two projections to extract processed scalar and vector tokens.

\subsection{Algorithm}

We detail the SE(3)-Hyena algorithm in Algorithm~\ref{alg:se3hyena}. The input projection, gating, and output projection operate in parallel for each token and thus of $O(N)$ complexity. The scalar and vector long convolutions are implemented with FFT and thus come with a complexity of $O(N log_2 N)$.

\begin{algorithm}[h!]
    \caption{SE(3)-Hyena forward pass}
    \label{alg:se3hyena}
    \begin{algorithmic}

    \REQUIRE $N$ input tokens $\mathbf{x} \in \mathbb{R}^{N \times 3}, \mathbf{f} \in \mathbb{R}^{N \times S}$ \\ 

    \vspace{2mm}
    \STATE \textcolor{lightgray}{\texttt{1. input projection:}}\\
    $[\mathbf{q}^{eqv}, \mathbf{k}^{eqv}, \mathbf{v}^{eqv}], [\mathbf{q}^{inv}, \mathbf{k}^{inv}, \mathbf{v}^{inv}] = \phi \left(\mathbf{x}, \mathbf{f}\right)$
    
    \vspace{2mm}
    \STATE \textcolor{lightgray}{\texttt{2. global context aggregation:}}\\
    $\mathbf{u}^{eqv} = \mathbf{q}^{eqv} \circledast_{\times} \mathbf{k}^{eqv}$ \\
    $\mathbf{u}^{inv} = \mathbf{q}^{inv} \circledast \mathbf{k}^{inv}$ \\

    \vspace{2mm}
    \STATE \textcolor{lightgray}{\texttt{3. gating:}}\\
    $m^{eqv}, m^{inv} = \gamma(\mathbf{u}^{eqv}, \mathbf{u}^{inv})$ \\
    $\hat{\mathbf{u}}^{eqv} = \sigma(m^{eqv}) \cdot \mathbf{u}^{eqv}$ \\
    $\hat{\mathbf{u}}^{inv} = \sigma(m^{inv}) \cdot \mathbf{u}^{inv}$ \\
    
    \vspace{2mm}
    \STATE \textcolor{lightgray}{\texttt{4. residual and output projection:}}\\
    \textcolor{mygray}{\textit{\# the cross product is between each $\hat{\mathbf{h}}^{eqv}_{i}$ and $\hat{\mathbf{v}}^{eqv}_{i}$}}\\
    $\hat{\mathbf{x}} = \xi(\mathbf{x} + \hat{\mathbf{u}}^{eqv} \times \mathbf{v}^{eqv})$ \\
    $\hat{\mathbf{f}} = \xi(\mathbf{f} + \hat{\mathbf{u}}^{inv} \cdot \mathbf{v}^{inv})$ \\
    
    \vspace{2mm}
    \STATE \textbf{Return:} $\hat{\mathbf{x}} \in \mathbb{R}^{N \times 3}, \hat{\mathbf{f}} \in \mathbb{R}^{N \times D}$
    
    \end{algorithmic}
\end{algorithm}

\begin{figure*}[t!]
  \centering
    \includegraphics[width=0.98\linewidth]{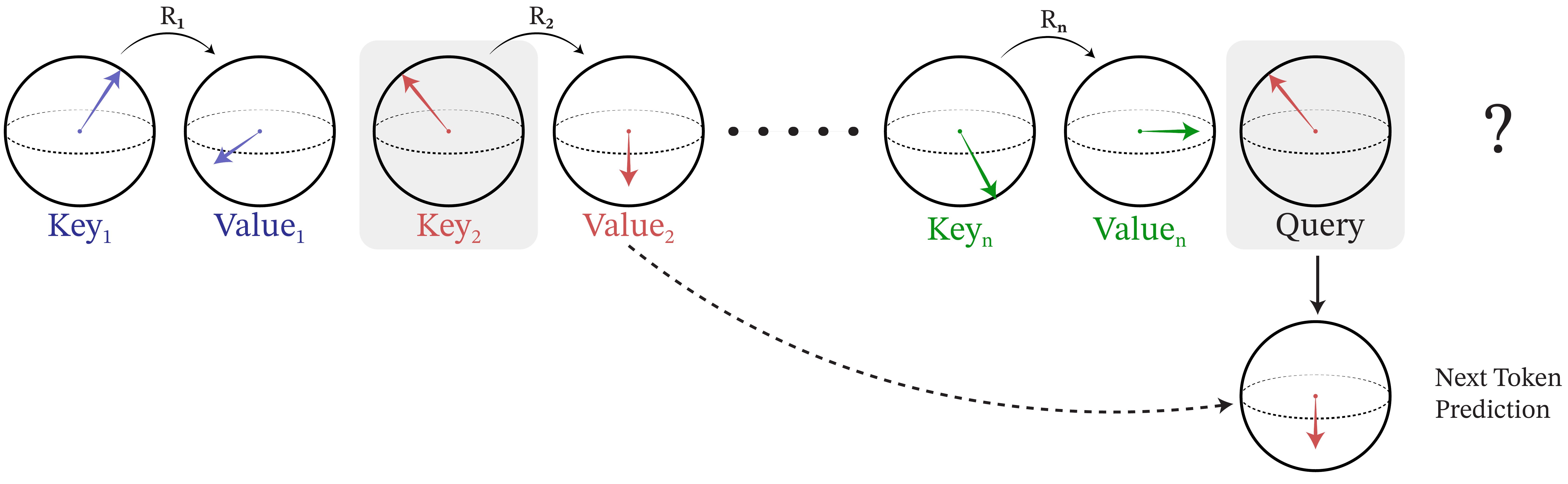}
    \caption{\small
    \textbf{Equivariant associative recall task.}
    An equivariant associative recall requires retrieving a vector token for a given vector query based on the context. The retrieval mechanism requires equivariance to rotation of tokens in a sequence. As standard associative recall serves to test the capability of models to learn global context, the equivariant associative recall task serves to test capability of models to learn global context with equivariance.
    }
  \label{fig:arec_task}
\end{figure*}

\section{Experiments}

\paragraph{Overview}
We conduct experiments to assess the performance of the SE(3)-equivariant Hyena in modeling global geometric context. We start with the associative recall (induction heads) task from mechanistic interpretability~\cite{elhage2021mathematical} that has become a standard for comparing efficiency of models in processing global context. Due to the equivariance requirement of geometric context, we have developed a vector-valued extension of the standard associative recall with equivariance to rotations. Next, we evaluate our model on the n-body dynamical system task, where the focus is on accurately predicting the positions of particles based on their initial velocities and coordinates while maintaining equivariance to rotations. Lastly, we compare the runtime and memory profiles of the equivariant Hyena against the equivariant transformer to highlight the computational efficiency of our method for long sequences.

\paragraph{Baselines} As a baseline, we evaluate the SE(3)-Hyena against the equivariant transformer, focusing specifically on global context modeling. To facilitate fair comparison, we aim to minimize architectural differences between the two models, altering only the global context mechanism. In the SE(3)-Hyena, we utilize scalar and vector long convolutions, whereas the transformer employs equivariant vector self-attention. The details on equivariant vector self-attention are in Appendix~\ref{sec:appendix_method}. We also include comparisons with non-equivariant versions of both the Hyena and transformer models to underscore the benefits provided by equivariance.

\begin{figure}[t]
    \centering
    \begin{minipage}{0.9\linewidth}
        \centering
        \includegraphics[width=\linewidth]{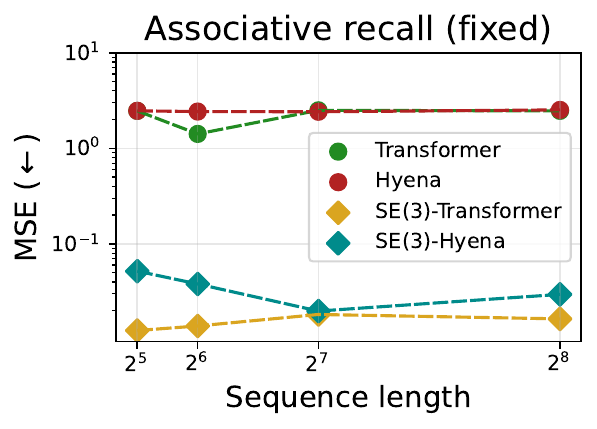}
    \end{minipage}\\
    \begin{minipage}{0.9\linewidth}
        \centering
        \includegraphics[width=\linewidth]{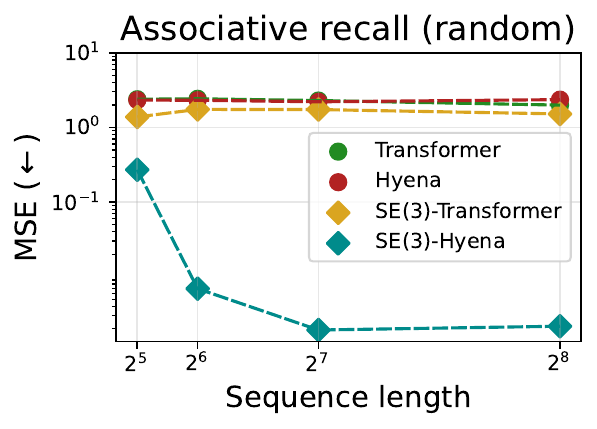}
    \end{minipage}
    \caption{\small \textbf{Top row:} The MSE between retrieved and target vectors for the fixed vocabulary associative recall task is plotted across various sequence lengths. Equivariant models effectively learn and generalize the underlying vocabulary across different orientations. \textbf{Bottom row:} The MSE for the random vocabulary associative recall task. The SE(3)-Hyena excels in learning the equivariant retrieval function, successfully associating target queries with their corresponding value vectors.}
    \label{fig:arec_result_fixed}
\end{figure}

\subsection{Equivariant associative recall}
\label{subsec: equivariant_associative_recall}

Associative recall~\cite{olsson2022context} is one of the standard mechanistic interpretability tasks used for quantifying the contextual learning capabilities of sequence models. In this task, a model is required to perform associative recall and copying; for instance, if a model previously encounters the bigram "Harry Potter" in a sequence, it should accurately predict "Potter" the next time "Harry" appears, by referencing its past occurrence~\cite{gu2023mamba}.

To adapt this for geometric contexts, we modify the standard associative recall to accommodate 3D vectors. In this version, the tokens within a bigram (key and value) relate to each other by a rotation matrix. The model processes a sequence of $N$ vector tokens concluding with a query token\footnote{key,value and query tokens here refer to associative recall task and different from keys,values and queries outputted by the input projection} and must predict a vector associated with this query seen earlier in the sequence, as illustrated in Figure~\ref{fig:arec_task}. The task ensures that rotating the entire sequence affects only the orientation of the predicted vector, not the key-value relationship within a bigram, thereby making the task equivariant to rotations. The task complexity depends on the number of tokens in a sequence and on vocabulary size where the vocabulary items are unique key-value vector bigrams.

We propose two versions of the equivariant associative recall task: random and fixed vocabulary versions. In the random vocabulary version, sequences are sampled from a randomly generated vocabulary at each training iteration while validation and test sets are fixed. Thus, an only way for a model to solve this task is to learn an equivariant retrieval mechanism that can associate a given query with a corresponding value vector. In the fixed vocabulary version, the vocabulary is fixed and shared among training, test and validation sets, and the sequences are randomly rotated during the training. This way, the fixed version tests the model's ability to learn underlying vocabulary and generalize it for various orientations.

\subsubsection{Implementation details} 

\paragraph{Dataset} We set the training data size to $2600$, and validation and test set size to $200$ sequences each. The vocabulary size is set to $4$ for both random and fixed versions. The vocabularies are generated as follows: one key-value bigram consists of two vectors with orientations sampled as random unit 3D vectors from an isotropic normal distribution, and with magnitudes randomly sampled from a uniform distribution in a range of $[1,\texttt{vocab\_size}]$. We generate datasets with various sequence lengths from $2^5$ to $2^8$ tokens. When generating sequences from a vocabulary, the last token is a key that corresponds to a target value. We additionally constraint the generation so the target key-value pair is present in a sequence at least once.

\paragraph{Models} The vector tokens are used as the input to the equivariant branch, while for the invariant branch, we use positional encoding features~\cite{vaswani2017attention} of dimension $16$ as the input. The SE(3)-Hyena model consists of $3$ SE(3)-Hyena operators followed by a mean pooling of equivariant features and one output equivariant MLP. In SE(3)-Hyena operator we set the hidden dimension to $16$ for invariant and to $128$ for equivariant streams. In the gating operator, we use a smaller dimension of $8$ for both equivariant and invariant features. This ends up in approximately $800k$ trainable parameters. Similarly, the equivariant transformer consists of $3$ SE(3)-equivariant vector self-attention blocks followed by a mean pooling and one equivariant MLP at the end. The hidden dimension is kept similar to SE(3)-Hyena which results in a nearly identical number of trainable parameters. Non-equivariant baselines consist of $3$ Hyena or standard self-attention blocks but with $3$ times larger hidden dimensions to balance the number of trainable parameters with equivariant models.

\paragraph{Training} We train all models for $300$ epochs with a batch size of $32$. We employ Adam optimizer~\cite{kingma2014adam} with an initial learning rate of $0.001$ and cosine learning rate annealing~\cite{loshchilov2016sgdr}. The weight decay is set to $0.00001$. Mean squared error is used as a loss function. For the fixed vocabulary experiment, we apply random rotation augmentation on sequences in the training batch, for the random vocabulary variant this is not necessary as sequences already appear in arbitrary orientations. Final models are selected based on the best validation loss.

\subsubsection{Results} The results for various sequence lengths are presented in Figure~\ref{fig:arec_result_fixed}. We record the mean squared error between predicted and ground truth vectors as a performance measure. 

For a fixed vocabulary variant, SE(3)-Hyena performs on par with the equivariant transformer across the whole range of sequence lengths. This demonstrates that both equivariant Hyena and transformer models can learn an underlying vocabulary and can generalize it for various orientations. We also observed that non-equivariant models were only able to learn the expectation across the training dataset.

For the random vocabulary variant, we observed that SE(3)-Hyena successfully learns the retrieval function to associate a target query with a corresponding target value vector. We observed better generalization for longer sequences which can be attributed to a higher frequency of target key-value bigram occurrence. Other models struggled to converge, with SE(3)-transformer only slightly outperforming non-equivariant models that only learn the expectation.

\subsection{N-body problem}
\label{subsec: dynamical_system_modeling}

In the field of dynamical systems, comprehending the time evolution of point sets within geometric spaces is crucial for a variety of applications, such as control systems~\cite{chua2018deep}, model-based dynamics in reinforcement learning~\cite{nagabandi2018neural}, and simulations of physical systems~\cite{watters2017visual}. This task inherently requires equivariance, as any rotations and translations applied to the initial set of particles must be consistently represented throughout their entire trajectory.

We benchmark SE(3)-Hyena on forecasting trajectories in the dynamical system where the dynamics is governed by physical interaction between $5$ charged particles that carry a positive or negative charge. Particles have a position and a velocity in 3-dimensional space. The objective is to predict the future positions of particles given initial positions and velocities. 

\subsubsection{Implementation details}

\paragraph{Dataset} We utilize the standard n-body benchmark from \citet{fuchs2020se,satorras2021n}. We sample $1000$ particles for training, $2000$ for validation, and $2000$ for testing. A training sample consists of equivariant features, i.e. initial positions $\mathbf{p}_{0} \in \mathbb{R}^{5 \times 3}$ and velocities positions $\mathbf{v}_{0} \in \mathbb{R}^{5 \times 3}$. Positions of the particles $\mathbf{p}_{1000} \in \mathbb{R}^{5 \times 3}$ after $1.000$ timesteps are used as labels.

\vspace{-3mm}
\paragraph{Models and training} The initial positions and velocities serve as the input to the equivariant branch, and one-hot encoded charges are used as the input to the invariant branch. The SE(3)-Hyena model consists of $2$ SE(3)-Hyena operators followed by one output equivariant MLP. The equivariant transformer baseline consists of $2$ vector self-attention blocks and one output equivariant MLP. For equivariant models, the invariant hidden dimension is set to $8$ and the equivariant hidden dimension to $16$. The gating in SE(3)-Hyena operator uses a hidden dimension of $8$ for both equivariant and invariant features. Non-equivariant baselines also consist of $2$ Hyena or standard self-attention blocks but with $3$ times larger invariant and equivariant hidden dimensions to match the number of trainable parameters with equivariant models. We also use the linear baseline model which is simply a linear motion equitation $\mathbf{p}_{t} = \mathbf{p}_{0} + t \cdot \mathbf{v}_{0}$. The models are trained for $300$ epochs with a batch size of $100$. We use Adam optimizer with a learning rate set to $0.0001$ and weight decay of $0.00001$. The mean squared error is used as the loss function.

\subsubsection{Results} The results are presented in Table \ref{tab:nbody}. The SE(3)-Hyena model outperforms non-equivariant baselines and performs on par with the equivariant transformer. Interestingly, the non-equivariant Hyena also slightly outperformed the non-equivariant transformer. This may stem from Hyena's structural similarities to SSMs~\cite{poli2023hyena} which are derived from the state-space representation of differential equation and hence are well-suited for modeling dynamical systems~\cite{hinrichsen2005mathematical}.

\begin{table}[t!]
\centering
\begin{tabular}{@{}llcl@{}}
\toprule
 & Method            & MSE    &  \\ \midrule
 & Linear            & 0.0322 &  \\
 & Transformer       & 0.0163 &  \\
 & Hyena             & 0.0150 &  \\
 & SE(3)-Transformer & \textbf{0.0019} &  \\
 & SE(3)-Hyena       & \textbf{0.0018} &  \\ \bottomrule
\end{tabular}
\caption{\small Mean squared error for the future position estimation in the N-body system experiment.}
\label{tab:nbody}
\end{table}

\subsection{Runtime and memory benchmarks}
\label{subsec:runtime}

We benchmark the runtime and memory consumption of a single layer SE(3)-Hyena against a single layer equivariant transformer when processing various sequence lengths. Similar to~\cite{dao2022flashattention,poli2023hyena}, we use random sequences for the runtime evaluation and we increase the sequence length until the SE(3)-Transformer runs out of memory. We record all runtimes on NVIDIA A10G GPU with CUDA version 12.2.

\begin{figure}[t!]
  \centering
    \begin{minipage}{0.9\linewidth}
        \centering
        \includegraphics[width=\linewidth]{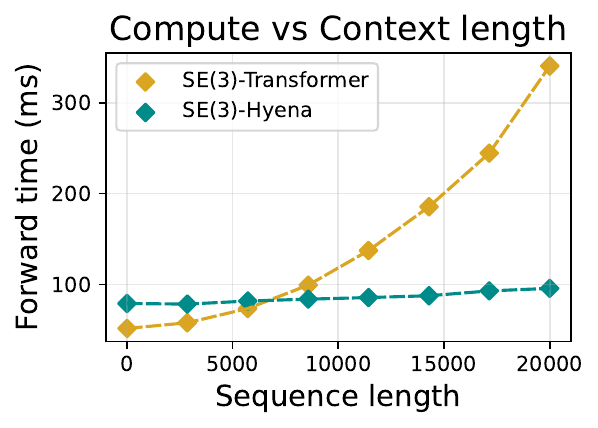}
    \end{minipage}\\
    \begin{minipage}{0.9\linewidth}
        \centering
        \includegraphics[width=\linewidth]{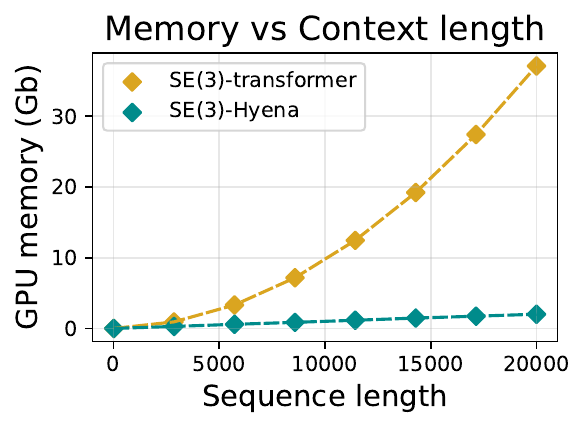}
    \end{minipage}

    \caption{\small \textbf{Top row:} Forward runtime comparison. SE(3)-Hyena scales sub-quadratically and achieves a considerable speedup compared to SE(3)-Transformer when processing long sequences. \textbf{Bottom row:} Total GPU memory utilization for equivariant Hyena and transformer models.}
  \label{fig:forward_time_comparison}
\end{figure}

The comparison is reported in Figure~\ref{fig:forward_time_comparison}, showing forward pass time in milliseconds and total GPU memory utilization in gigabytes. SE(3)-Hyena easily scales to longer sequences whereas SE(3)-Transformer is $3.5\times$ slower than our model for a sequence length of $20k$ tokens. Similarly to \citet{nguyen2024hyenadna}, we observed Hyena to be slightly slower for shorter sequences. This can be attributed to expensive I/O between layers of the GPU memory hierarchy and can be addressed with the recent GPU-optimized versions of the FFT long convolution~\cite{fu2023flashfftconv}. Regarding memory usage, for the sequences length of $20k$ tokens the equivariant Hyena requires $18$ times less GPU memory than transformer for a forward pass. Moreover, we observed that when the equivariant transformer runs out of memory on $>20k$ tokens, \textit{our model supports up to $3.5M$ tokens on a single GPU allowing for $175$ times longer geometric context}. This memory efficiency is attributed to the FFT long convolution that avoids materializing a quadratic self-attention matrix.

\section{Conclusions}
We introduced the SE(3)-Hyena operator being, to the best of our knowledge, the first equivariant long-convolutional model with sub-quadratic complexity for global geometric context. Through experiments on the dynamical system and novel equivariant associative recall task, we demonstrated that the equivariant long-convolutional model can perform competitively to the equivariant self-attention while requiring a fraction of the computational and memory cost of transformers for long context modeling. Our scalable equivariant model efficiently captures the global context, highlighting its potential for a multitude of future applications in various domains.

\paragraph{Limitations and Future work}
This work introduces a novel approach for equivariant modeling of global geometric context at scale, with initial experiments designed to validate the fundamental principles of our method. While these experiments confirm the key advantages of our approach, they represent the first step of a comprehensive experimental analysis that is necessary to uncover the model's capabilities across a wider range of real-world tasks. On a technical side, an interesting direction for future improvement is to adapt the vector convolution to function across arbitrary dimensions as it currently relies on a cross product, which is only feasible in $3$ and $7$ dimensions~\cite{massey1983cross}. Also, our method relies on the FFT convolution that is not permutation equivariant. While this is not a significant limitation in domains such as molecular systems, where canonical ordering is typically available~\cite{jochum1977canonical}, it becomes critical in fields like point cloud processing, where establishing a canonical order is challenging. Enhancing the long-convolutional framework to incorporate permutation equivariance or learning this geometric constraint from the data \cite{moskalev2023ginvariance} could offer substantial advantages in these areas.



\bibliography{bibliography}
\bibliographystyle{icml2024}

\newpage
\appendix
\onecolumn

\section{Equivariant vector long convolution}

We provide Pytorch implementation for the rotation-equivariant (without centering) vector long convolution in Code~\ref{lst:pytorch_conv}.

\begin{lstlisting}[caption={\small Pytorch implementation of the equivariant vector long convolution.}, label={lst:pytorch_conv}]

class VectorLongConv(nn.Module):
    def __init__(self):
        super(VectorLongConv, self).__init__()
                
        # L cross-prod tensor factorized:
        l_reduced = torch.FloatTensor([[1, 0, 0],
                                       [1, 0, 0],
                                       [0, 1, 0],
                                       [0, 1, 0],
                                       [0, 0, 1],
                                       [0, 0, 1]])
        self.register_buffer("l_reduced", l_reduced, persistent=False)
        
        # H cross-prod tensor factorized:
        h_reduced = torch.FloatTensor([[0 , 1, 0],
                                       [0 , 0,-1],
                                       [-1, 0, 0],
                                       [0 , 0, 1],
                                       [1 , 0, 0],
                                       [0 ,-1, 0]])
        self.register_buffer("h_reduced", h_reduced, persistent=False)
        
        # P cross-prod tensor factorized:
        p_reduced = torch.FloatTensor([[0,0,0,1,0,1],
                                       [0,1,0,0,1,0],
                                       [1,0,1,0,0,0]])
        self.register_buffer("p_reduced", p_reduced, persistent=False)
        
    def forward(self, q, k):

        # q,k: (batch, sequence, dim=3)
        B, N, D = q.shape        
                
        # batchify L,H,P reduced matrices
        l_reduced = self.l_reduced[None,None].repeat(B,N,1,1)
        h_reduced = self.h_reduced[None,None].repeat(B,N,1,1)
        p_reduced = self.p_reduced[None,None].repeat(B,N,1,1)
        
        # expand inputs with reduced L,H,P matricies
        q_expd = torch.matmul(l_reduced, q.unsqueeze(-1)).squeeze(-1)
        k_expd = torch.matmul(h_reduced, k.unsqueeze(-1)).squeeze(-1)
                
        # fft conv
        fft_q = torch.fft.rfft(q_expd, n=N, dim=1)
        fft_k = torch.fft.rfft(k_expd, n=N, dim=1)
        fft_conv_qk = torch.fft.irfft(fft_q*fft_k, n=N, dim=1)

        # reduce to vector product and normalize
        u = torch.matmul(p_reduced, fft_conv_qk.unsqueeze(-1)).squeeze(-1) / N
        
        return u
        
\end{lstlisting}

\section{Equivariant vector self-attention}
\label{sec:appendix_method}

Since SE(3)-Hyena utilizes long convolutions based on cross products, we design cross product equivariant self-attention to align the global geometric context aggregation for Hyena and transformer models. Firstly, the self-attention tensor is built from all pairs of cross products between queries and keys. Then, the $L_2$ norm matrix is extracted and softmax is run row-wise on top of this matrix. The resulting softmax matrix serves as a selection mechanism to select vectors from the cross product tensor. Finally, the values are cross-multiplied with the self-attention tensor. This yields a set of processed values modulated with global self-attention information. 

\vspace{-3mm}
\paragraph{Cross product vector self-attention} 
Consider sequences of $N$ vector queries, keys, and values denoted as $\mathbf{q},\mathbf{k},\mathbf{v} \in \mathbb{R}^{N \times 3}$. We construct a query-key cross product tensor $\mathbf{C} \in \mathbb{R}^{(N \times N) \times 3}$ where each element $\mathbf{C}_{ij} = \mathbf{q}_{i} \times \mathbf{k}_{j}$, or using Levi-Civita notation as in Eq.~\ref{eq:vec_conv_levi}, $\mathbf{C}_{ij}[l] = \varepsilon_{lhp} \mathbf{q}_{i}[h] \mathbf{k}_{j}[p]$. To integrate a softmax selection mechanism, as in standard self-attention, we first compute a matrix $\eta(\mathbf{C}) \in \mathbb{R}^{N \times N}$ containing the $L_2$ norms of all cross products, specifically $\eta(\mathbf{C})_{ij} = \| \mathbf{q}_{i} \times \mathbf{k}_{j} \|_2$. Applying softmax to $\eta(\mathbf{C})$ then determines the vector pairs to select from the cross product tensor. Lastly, the values $\mathbf{v}$ are cross-multiplied with the softmax-filtered cross product tensor. Overall, the equivariant vector self-attention reads as:

\begin{gather}
    \label{eq:sa_vec}
    \mathbf{S} = \texttt{softmax}(\frac{1}{\sqrt{N}}\eta(\mathbf{C})) \odot \mathbf{C} \\
    \label{eq:sa_ucross}
    \mathbf{u}_{i} = \frac{1}{N}\sum_{j=1}^{N} \mathbf{S}_{ij} \times \mathbf{v}_{j}
\end{gather}

where the softmax is applied row-wise, and $\odot$ stands for element-wise product. Consequently, $\mathbf{S} \in \mathbb{R}^{(N \times N) \times 3}$ represents a tensor that encapsulates a soft selection of cross products between $\mathbf{q}_{i}$ and $\mathbf{k}_{j}$. Initially, we considered using just $\texttt{softmax}(\frac{1}{\sqrt{N}}\eta(\mathbf{C}))$ as a self-attention matrix, but early experiments indicated that the method outlined in Eq.~\ref{eq:sa_vec} yields better results. Additionally, we found that normalizing the sum by $1/N$ in Eq.~\ref{eq:sa_ucross} further improves convergence.

Since the tensor $\mathbf{C}$ is constructed using cross products, it naturally maintains equivariance to rotations of queries and keys. Furthermore, the softmax is applied to the $L_2$ norms of these cross products, which makes it invariant to rotations. Consequently, the self-attention tensor $\mathbf{S}$ is a product of rotation-invariant scalar and rotation-equivariant vector quantities, rendering it rotation-equivariant. The Eq.~\ref{eq:sa_ucross} further preserves rotation-equivariance due to the inherent equivariance of the cross product. Equivariance to translations can be achieved by initially centering the data (subtracting the center of mass) and then re-centering the resulting tokens. 

We provide Pytorch implementation for the rotation-equivariant (without centering) vector self-attention in Code~\ref{lst:pytorch_sa}.

\begin{lstlisting}[caption={Pytorch implementation of equivariant vector self-attention.}, label={lst:pytorch_sa}]

class VectorSelfAttention(nn.Module):
    def __init__(self):
        super(VectorSelfAttention, self).__init__()
        
    def forward(self, q, k, v):

        # q,k,v: (batch, sequence, dim=3)
        B, N, D = q.shape
        
        # cross product matrix C: (B, N, N, 3)
        q_expd, k_expd = q.unsqueeze(2), k.unsqueeze(1)
        C = torch.cross(q_expd, k_expd, dim=-1)
        
        # \eta(C) matrix of norms and softmax: (B, N, N, 1)
        eta_C = C.norm(dim=-1, keepdim=True)
        sm_eta_C = nn.functional.softmax(eta_C/(N**0.5), dim=2)
        
        # S matrix: (B, N, N, 3)
        S = sm_eta_C*C
        
        # compute u = S x v: (B, N, 3)
        pre_u = torch.cross(S, v.unsqueeze(2), dim=-1)
        u = pre_u.mean(dim=2)
        
        return u

\end{lstlisting}





\end{document}